\documentclass[runningheads]{llncs}
\usepackage{graphicx}
\usepackage{color} 
\usepackage{comment} 
\usepackage{amsmath}
\usepackage{algorithm}
\usepackage[noend]{algorithmic}
\usepackage[misc]{ifsym}

\begin{document}

\title{
Stable deep reinforcement learning method by predicting uncertainty in rewards as a subtask
}
\titlerunning{
Stable DRL method by predicting uncertainty in rewards as a subtask
}

\author{
Kanata Suzuki\inst{1,2} \and 
Tetsuya Ogata\inst{2,3}(\Letter)
}
\authorrunning{
K. Suzuki and T. Ogata
}

\institute{
$^{1}$ Artificial Intelligence Laboratories, Fujitsu Laboratories LTD., Kanagawa, Japan \\
\email{suzuki.kanata@jp.fujitsu.com} \\
$^{2}$ School of Fundamental Science and Engineering, Waseda University, Tokyo, Japan \\
$^{3}$ Artificial Intelligence Research Center, Tsukuba, Japan \\
\email{ogata@waseda.jp}
}

\maketitle

\begin{abstract}
In recent years, a variety of tasks have been accomplished by deep reinforcement learning (DRL). 
However, when applying DRL to tasks in a real-world environment, designing an appropriate reward is difficult. 
Rewards obtained via actual hardware sensors may include noise, misinterpretation, or failed observations. 
The learning instability caused by these unstable signals is a problem that remains to be solved in DRL. 
In this work, we propose an approach that extends existing DRL models by adding a subtask to directly estimate the variance contained in the reward signal. 
The model then takes the feature map learned by the subtask in a critic network and sends it to the actor network. 
This enables stable learning that is robust to the effects of potential noise. 
The results of experiments in the Atari game domain with unstable reward signals show that our method stabilizes training convergence. 
We also discuss the extensibility of the model by visualizing feature maps. This approach has the potential to make DRL more practical for use in noisy, real-world scenarios.

\end{abstract}

\section{Introduction}
Although deep reinforcement learning (DRL) has been shown to have high performance in various fields, some challenges remain regarding the stability of training.
In applications such as games~\cite{dqn,alpha-go}, by designing the score as a reward value, it is possible to obtain a model that obtains a performance comparable to that of a human.
However, many DRL studies~\cite{levine-1,levine-2} only conducted experiments with simple reward signals designed by the experimenter.
There is a gap between these scenarios and real environments, which often have unstable reward signals.
This is an essential issue for DRL because its performance is sensitive to reward design.
Therefore, a learning method that is robust to noise in the reward signals is needed.

Noise in the DRL reward function can occur for several reasons;
a typical example is the errors that occur during observation.
In the real-world, reward functions are not perfect.
The rewards derived from the actual environment via hardware sensors may include noise, misinterpretations, and observation failures.
When misinterpretations or observation failures occur, the reward value may be calculated as an entirely different value.
Another case in which noise may occur is the use of feature values as signals for training a neural network.
A deep neural network extracts low-dimensional feature vectors from high-dimensional sensor information.
Furthermore, we can use extracted features as the target signals of another network.
For example, some studies use the feature values of images of target values as a signal to optimize robot behaviors~\cite{Grasp2Vec,sii20}.
When employing a reward signal to acquire advanced behavior, a variety of noise types not intended by the experimenter will occur.

\begin{figure}[t]
	\begin{centering}
        \includegraphics[width=10.0cm]{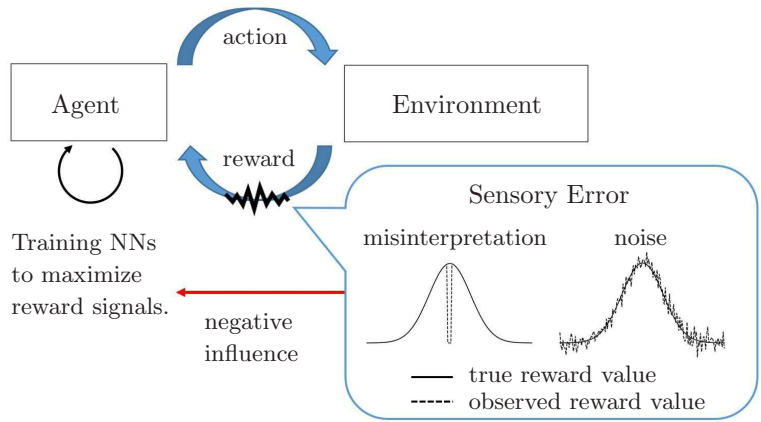}
        \put(-268,117){\normalsize Agent}
        \put(-139,117){\normalsize Environment}
        \put(-203,143){\small action}
        \put(-203,97){\small reward}
        \put(-283,65){\small Training NNs}
        \put(-283,53){\small to maximize }
        \put(-283,41){\small reward signals.}
        \put(-210,35){\small negative}
        \put(-210,23){\small influence}
        \put(-110,15){\small true reward value}
        \put(-110,5){\small observed reward value}
        \put(-110,83){\normalsize Sensory Error}
        \put(-150,67){\small misinterpretation}
        \put(-55,67){\small noise}
        \caption{
There are several sources of noise in the reward of DRL in a real-world environment.
If noisy signals are used as target signals, they delay convergence in training.
        }
	\end{centering}
\end{figure}

Among the types of unstable reward signals listed above, in this study, we focus on fine-grained noise in the signals, 
which has been referred to as ``sensory error'' in previous research~\cite{tom2017}.
In this case, we assume that the reward value is a continuous value rather than a binary signal.
These unstable reward signals may inhibit DRL training; for instance, delaying convergence during training.
Therefore, a DRL model needs a learning method that considers the uncertainty of the signals and updates its parameters appropriately.

Hence, we propose a stable learning method for DRL with unstable reward signals by directly estimating the variance of the rewards and adjusting the parameter update amount.
We incorporate an estimation of the variance of the target signals into the model as a subtask.
This makes it possible to extend the original model without significant changes to its configuration.
In addition, we use an attention branch network (ABN)~\cite{abn} structure that incorporates the feature map of the subtasks into the main task.
This conveys the learning results of the subtask to the policy network.
We evaluated our method on the Atari game domain in the Open AI Gym~\cite{gym}.
To verify the model can stabilize training convergence in an environment where rewards are unstable, we conducted experiments in which we added artificial noise to the reward signals. The results show that our extension to the base model improves performance.
The primary contributions of this study are as follows.
\\ 1) A model is proposed to stabilize training convergence that incorporates a subtask that estimates the variance in the rewards signals.
\\ 2) An evaluation of the performance of models trained with disturbed rewards shows that the proposed approach improves performance.

\section{Related Works}
Several methods have been proposed to improve the convergence and stability of DRL, and they fall into two main types.
One approach optimizes training, whereas the other one reduces variance.
The method proposed in this study uses the latter approach.

To increase the convergence and stability of DRL, many optimization methods have been proposed.
RMSprop~\cite{rmsprop} is an approach based on AdaGrad~\cite{adagrad}, which adjusts the learning rate with respect to the frequency of each parameter update and results in a rapidly decreasing learning rate.
Adam~\cite{adam} is a further improvement on traditional optimization methods and is used in many studies on deep learning.
Furthermore, in terms of variance control, methods such as SAG~\cite{sag}, SDCA~\cite{sdca} and stochastic variance reduction (SVR)~\cite{svr} have also been proposed.
SVR-DQN~\cite{svr-dqn} reduces the variance resulting from approximate gradient estimation.
These optimizers are essential advances in the convergence and stability of DRL.
However, in many studies, experimenters empirically use what is appropriate for the model.

Several previous studies consider the uncertainty of the target signals.
The authors of \cite{tom2017} define a Markov decision process in the presence of misinterpretations of the rewards and observation failures and propose a way to deal with rewards that are not correct through sampling.
However, the study remains an investigation of table-style rewards and does not consider a continuous control method.
There is also a study that addresses the overestimation error in Q-learning.
Double DQN uses separate Q-networks for action selection and Q-function value calculation~\cite{double-dqn}.
The model is able to deal with overestimation errors that replace positive bias with negative bias.
Another approach is to reduce the target approximation error.
An efficient way to reduce this is to use the average of multiple models.
The average DQN estimates the current action value using the previously computed K-value~\cite{average-dqn}.
In contrast, an estimator has been proposed to reduce the variance of the reward signal~\cite{noise-2}.
This model updates the discount value function instead of the sampled rewards.

In this study, we propose a model extension that uses a mechanism to estimate the variance of the reward signals directly.
The model estimates the mean and variance of the rewards obtained from the environment, and
it can be easily integrated with the base model as a subtask.
Using the estimated variance, the model updates its parameters to reduce the effects of noise.
To apply the above approach, we adopt an actor-critic type network that predicts the policy and state value using the actor and critic, respectively.

\section{Method}
To solve the problem described above, we extend the DRL model to solve subtasks that predict the variance of reward signals.
Figure 2 shows an overview of the proposed network architecture. 
The model consists of a base neural network model and an extended branch network.
We describe the base model in Section 3.1 and the proposed extension in Section 3.2.

\subsection{Base model: ABN-A3C}
As the base model, we adopt a DRL model that combines the ABN~\cite{abn} and asynchronous advantage actor critic (A3C)~\cite{a3c}.
We choose an actor-critic type network so that we may incorporate a subtask that predicts the variance of the reward signals.
Moreover, the ABN enables us to visualize the focus of the subtask using a feature map.
The base model, ABN-A3C, consists of a feature extractor that extracts features from the input image, a value branch that outputs state values, and a policy branch that outputs actions.
The policy branch also uses the feature map $f(s_t)$ of the value branch as input.
Feature map $f(s_t)$ is extracted from the current state $s_t$, and the value branch outputs the maximum value of feature map $f(s_t)$ using global max pooling.
This emphasizes the more distinctive pixels in the feature map of a subtask when it is incorporated with the main task.
The details of each model are described below.

\textbf{A3C: }
The training of an A3C is stabilized by using various policy searches while running multiple agents in parallel.
In asynchronous learning in multiple environments, there is a globally shared parameter and a separate parameter for each thread.
Each worker's parameters are copied from the global network's parameters.
The parameters learned by agents under different environments are reflected asynchronously in the global network.
The gradient exponential moving average of RMSprop, which is used as the optimizer, is also shared globally.

A3C takes advantage of its ability to train online and updates the state value using an estimation of the reward several steps in the future as opposed to a method that estimates the reward in the next step only. As a result, learning is more stable because a more likely estimation error of the current state value is used. 
The value of $adv$, which is used to update the estimated value, is calculated by the following equation. 
\begin{eqnarray}
adv = \sum_{i=0}^{k-1}{\gamma^i r_{t+1}} + \gamma^k V(s_{t+k}) - V(s_t)
\end{eqnarray}
where $k$ indicates how many future steps are used in the prediction.
We decided on the prediction step that gave the best results by trying some experimental settings.
In our experiment, we set the prediction step $k=5$.

The A3C trains two models: an actor network, which represents the behavior of the agent, and a critic network, which predicts the expected rewards. 
An actor network is trained to predict the probability of taking action in a certain state $\pi$. 
The critic network is trained to predict the estimated value of state $V$. 
Because the estimated values are independent, they are easy to learn even when the action is continuous.

\textbf{ABN: }
An ABN is a model that makes it possible to visualize the areas of focus and improve the accuracy of the network by incorporating feature maps of the subtask into the main task.

In ABN, we compute a new feature map $g'(s_t)$ from the feature map $f(s_t)$ of the value branch and the output of the feature extractor using the following residual mechanism~\cite{res}:
\begin{eqnarray}
g'(s_t) = (1+f(s_t))*g(s_t)
\end{eqnarray}
The state value in the current state is reflected in the action, and the loss of the original feature map is suppressed.
The action is predicted by inputting $g'(s_t)$ to the LSTM network of the policy branch.
Here, feature map $f(s_t)$ represents the features for optimizing a subtask.
By visualizing the feature map overlaid on the input image, it is possible to show where the network is focusing its attention in the input image.

\begin{figure}
    \includegraphics[width=12.2cm]{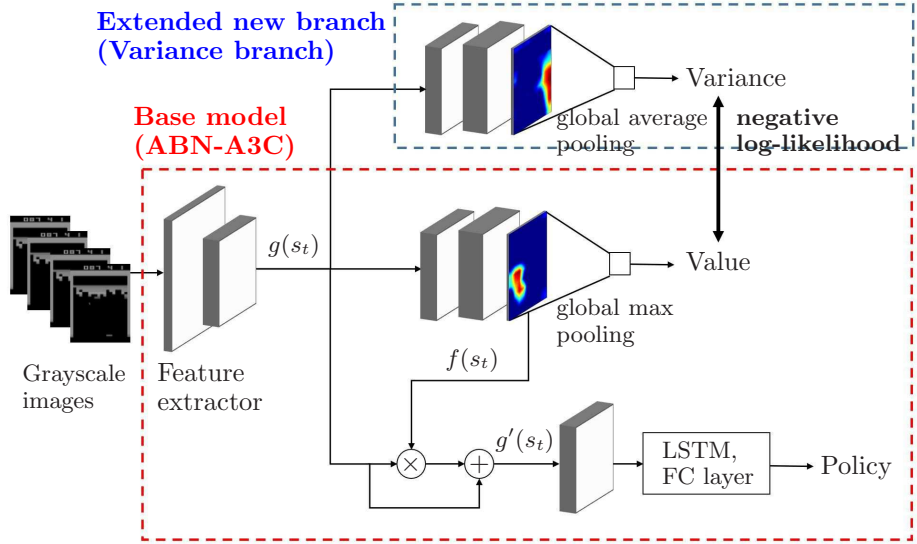}
    \put(-89,175){\normalsize Variance}
    \put(-89,105){\normalsize Value}
    \put(-38,28){\normalsize Policy}
    \put(-289,63){\normalsize Feature}
    \put(-289,52){\normalsize extractor}
    \put(-340,63){\small Grayscale}
    \put(-340,53){\small images}
    \put(-247,112){\small $g(s_t)$}
    \put(-180,69){\small $f(s_t)$}
    \put(-161,38){\small $g'(s_t)$}
    \put(-171,29){\normalsize $+$}
    \put(-197,29){\normalsize $\times$}
    \put(-138,88){\small global max}
    \put(-138,78){\small pooling}
    \put(-138,160){\small global average}
    \put(-138,150){\small pooling}
    \put(-98,34){\small LSTM,}
    \put(-98,24){\small FC layer}
    \put(-312,196){\normalsize \textcolor{blue}{\textbf{Extended new branch}}}
    \put(-312,185){\normalsize \textcolor{blue}{\textbf{(Variance branch)}}}
    \put(-298,160){\normalsize \textcolor{red}{\textbf{Base model}}}
    \put(-298,149){\normalsize \textcolor{red}{\textbf{(ABN-A3C)}}}
    \put(-70, 160){\small \textbf{negative}}
    \put(-70, 150){\small \textbf{log-likelihood}}
	\caption{
Overview of the proposed network for predicting the uncertainty of signals as a subtask.
The red frame indicates the base model, and the blue frame indicates the proposed extended branch. 
	}
	\label{fig1}
\end{figure}

\subsection{Variance branch for predicting uncertainty in rewards}
To stabilize the learning convergence, we extend the base model described in the previous section.
The aim is to optimize the model's parameters while ignoring the effects of reward noise.
Here, the reward signal with noise is assumed to have been generated according to some probability distribution from an unknown generative model.
We assume a Gaussian distribution in this study.
We use the branching structure of ABN to add a new branch called the variance branch, which takes the feature map as input.
The variance branch is similar to a stochastic multi-time scale recurrent neural network (SMTRNN)~\cite{smtrnn}.

The SMTRNN is a type of recurrent neural network that enables the predictive learning of probability distributions based on likelihood maximization.
The model extends the conventional learning method of point prediction based on squared-error minimization.
It learns to minimize the negative log-likelihood and obtains the stochastic structure that underlies the target signals.
To estimate the variance of the prediction of state value $V^{\pi}(s_t)$, the probability density $p(r_t|s_t,\theta)$ of reward $r_t$ at time step $t$ during an episode is deﬁned as
\begin{eqnarray}
p(r_t | s_t,\theta)=\cfrac{1}{\sqrt{2\pi\nu_t}}exp \left( -\cfrac{(V^{\pi}(s_t)-r_t)^2}{2\nu_t} \right)
\end{eqnarray}
and the log-likelihood L is defined as
\begin{eqnarray}
L = \prod_{t=1}^T p(r_t | s_t,\theta)
\end{eqnarray}
where $\theta$ denotes the model parameters.
This process is equivalent to minimizing the weighted prediction error by dividing the output error by predicted variance $v$.
The model learns while ignoring errors in rewards that contain large variance, i.e., large noise.
As a result, the training for the state value is stabilized.

In our method, the squared-error calculation of the state value and reward is replaced by the above function.
In addition, the configuration of the variance branch is based on the value branch.
To smooth the entire feature map of the variance branch, we adopt global average pooling in the final layer.
The stabilization of the state value prediction caused by the variance prediction is reflected in the stability of the behavioral prediction in the policy branch.
This is a result of the incorporation of the feature-map mechanism of ABN described in the previous section.
Visualization of the added branch network is also possible.

\section{Experiments}
\subsection{Model and environment setup}
To evaluate our method, we used the model to learn to play the Atari games in the Open AI Gym~\cite{gym}.
We used three games: Break Out, Sea Quest, and Pong.
As the input image for each game, we used $84\times84$ grayscale images of four time steps, and
for the training, we used RMSprop~\cite{rmsprop} as the optimizer.
Its learning rate is $7\times10^{-4}$, and the discount rate is 0.99.
The number of workers in the A3C is 32.
Table I lists the parameters of our model.
We determined the above parameters by trial-and-error, choosing the parameter set that yielded the best results.

\begin{table}
  \centering
  \begin{tabular}{c|c}
    \multicolumn{2}{c}{TABLE I: Structures of the networks} \\
    \hline
    Network         & Dimensions \\
    \hline \hline
    Feature Extractor & conv@16chs - BN - Relu -  \\ 
                      & conv@32chs - BN - Relu  \\ 
    \hline
    Value Branch & conv@32chs - BN - Relu -  \\ 
                 & conv@64chs - BN - Relu -  \\ 
                 & conv@1chs - BN - MaxPooling  \\ 
    \hline
    Policy Branch & eq.(2) - conv@32chs - BN - Relu -  \\ 
                  & LSTM@256 - FC@ActionNum  \\ 
    \hline
    Variance Branch & conv@32chs - BN - Relu -  \\ 
                 & conv@64chs - BN - Relu -  \\ 
                 & conv@1chs - BN - AvePooling -exp  \\ 
    \hline 
  \end{tabular}
  \begin{center}
  BN: Barch normalizaion, FC: Fully connect
  \end{center}
\end{table}

\subsection{Evaluation metrics}
To evaluate the effectiveness of the proposed method, we added artificial noise to the reward signals in the experiments.
The noise followed a Gaussian distribution of variance $\sigma^2$.
In our experiments, we set $\sigma^2$ to 0.0, 0.03, and 0.05. 
When $\sigma^2=0.0$, there is no noise in the reward signals, and 
the noise increases as $\sigma$ increases.
We compared the proposed method with a base model that does not have a mechanism for estimating the variance in the reward signals.
We performed experiments in each game environment five times while changing the initial weights of the model.

\section{Results and Discussion}
In this section, we present the results of experiments in multiple game environments to evaluate the robustness of the proposed method to noise in the reward signals. 
We also present a feature map of the model and analyze the points of focus in each game. 
Finally, we discuss the suitability of the proposed method for other deep neural network models.

\subsection{Atari game performance}
The scores during the training of the proposed and base models for each game are shown in Fig. 3.
The results show that the variance in the results increases with the variance in the reward signals.
The time to convergence increases because the teaching signal given to the model is not stable.
The proposed method converges faster than the ABN-A3C base model, regardless of the size of the variance.
However, the maximum score of the proposed method is about the same as that of the base model.
These results show that the proposed method predicts the mean of the reward signal and converges to the same results as the base model in less time.

\begin{figure}[t]
	\begin{centering}
        \includegraphics[width=12.2cm]{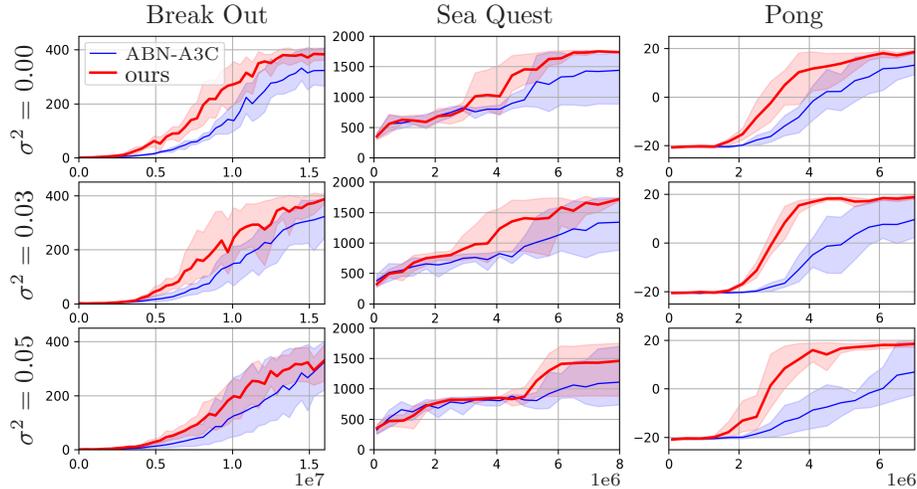}
        \put(-294,170){\normalsize Break Out}
        \put(-184,170){\normalsize Sea Quest}
        \put(-60, 170){\normalsize Pong}
        \put(-345,120){\normalsize \rotatebox{90}{$\sigma^2=0.00$}}
        \put(-345,65){\normalsize \rotatebox{90}{$\sigma^2=0.03$}}
        \put(-345,10){\normalsize \rotatebox{90}{$\sigma^2=0.05$}}
        \put(-302,156){\scriptsize ABN-A3C}
        \put(-302,148){\footnotesize ours}
        \put(-14, -5){\scriptsize 1e6}
        \put(-126, -5){\scriptsize 1e6}
        \put(-238, -5){\scriptsize 1e7}
        \caption{
Change in score during training under each condition (i.e., the type of game and level of reward signal noise).
The vertical axis of each figure shows the score, and the horizontal axis shows the total number of worker epochs.
Each color area shows the maximum and minimum range of the score. 
The red lines indicate the results of our method, and the blue lines indicate the results of the ABN-A3C base model.
        }
	\end{centering}
\end{figure}

The proposed method converges faster than the base model in all games, regardless of the level of noise in the reward signal.
This is also true when there is no noise ($\sigma^2=0.0$).
Although rewards are given discrete values in the standard games in the Atari game domain, the results suggest that predicting the mean of the rewards is an effective strategy.
We think that was because atari's ordinary rewards include uncertainty. 
For example, in atari games, not all rewards given are valid. 
Our method may learn to ignore temporary rewards that cannot maximize cumulative rewards.

When the levels of noise are low ($\sigma^2=0.03$), the results of the proposed and base model differ the most.
The final convergence score of the base model varies depending on the initial weight values, which are randomly chosen.
In contrast, the performance of the proposed method does not depend on these values.
The proposed method adequately learns the variance in the rewards and stabilizes the training of the policy network.
However, when the level of reward noise increases ($\sigma^2=0.05$), the results of the proposed method are also worse.
When the noise reaches a certain level, the training is substantially disturbed.
These results demonstrate the robustness of the proposed DRL method to unstable reward signals.
In our experiment, the noise added to the reward was artificially set; hence, the impact of realistic reward noise on learning needs to be considered in future work.

\subsection{Visualization of the feature map}
Next, we visualize the feature map of the proposed model to ensure that the model focuses on the appropriate areas of the feature map.
The feature map for each condition superimposed on the input image is shown in Fig. 4.
In Break Out, the feature map shows that the model focuses on the movement of the ball.
Furthermore, when the number of blocks decreases, the area of attention moves to the blank regions above the blocks (see the results for $\sigma^2=0.03$).
The variance branch's feature map is similarly active, focusing mainly on areas of significant change on the screen.
In Sea Quest, the feature map indicates focus on agents, enemies, and the bars representing the remaining oxygen.
There is less movement in the feature map than in Breakout, which may be because this is a game in which the agent employs a waiting strategy.
In contrast, in Pong, the feature map is not informative in most cases, even when the obtained scores reach their upper bounds.
Reviewing the gameplay after training, we found that the agent repeated a specific pattern of behavior to score points.
This may be because Pong itself does not require any complex behavior.

\begin{figure}
	\begin{centering}
        \includegraphics[width=12.2cm]{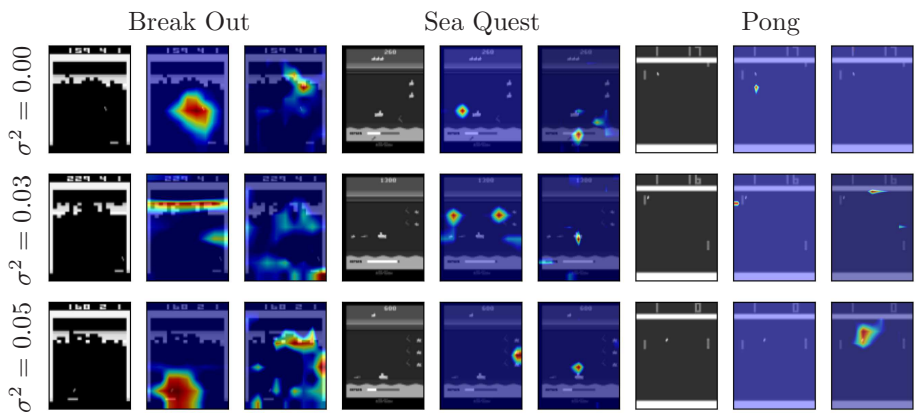}
        \put(-300,145){\normalsize Break Out}
        \put(-188,145){\normalsize Sea Quest}
        \put(-68, 145){\normalsize Pong}
        \put(-345,97){\normalsize \rotatebox{90}{$\sigma^2=0.00$}}
        \put(-345,48){\normalsize \rotatebox{90}{$\sigma^2=0.03$}}
        \put(-345,0){\normalsize \rotatebox{90}{$\sigma^2=0.05$}}
        \caption{
Examples of visualization of feature maps for each condition.
The input image, the feature map of the value branch, and the feature map of the variance branch are shown, respectively.
The value of each feature map is higher as it becomes red.
        }
	\end{centering}
\end{figure}

The above results confirm the effectiveness of our proposed method.
The model paid attention to the appropriate areas on the feature map, even in environments with unstable reward signals.
Furthermore, the regions of focus of the variance branch feature maps differ from those of the value branch feature maps.
In other words, each branch plays a different role in the network.

\subsection{Scalability}
The proposed method is broadly applicable to many conventional networks because it does not require significant changes to the configuration of the original model.
However, because the subtask for predicting variance requires the prediction of state values, the network to be extended should be an actor-critic type network.
Because the network is extended using a branch structure, the computational complexity of the network increases; however, parallel computation is possible.
Hence, the learning and prediction times should not be much different from those of the original network.

The combined method of variance prediction and feature-map visualization could be used in applications other than DRL.
We are investigating an extension to recurrent neural networks for end-to-end robot control~\cite{kase,suzuki}.
Robot control is a particularly promising application because it is often affected by real-world noise.

\section{Conclusion}
In this study, we proposed a stable reinforcement learning method for scenarios in which the reward signal contains noise. 
We incorporated a subtask into an actor-critic-based DRL method. 
The model directly estimates the variance included in the reward obtained from the environment. 
Moreover, we input the feature map learned by the subtask in the critic network to the actor network. 
We evaluated our method in the Atari game environment of the Open AI Gym.
Our method enables us to stabilize the convergence of learning in an environment in which rewards are unstable.
In future work, we plan to extend our method to real robot tasks.

\section*{Acknowledgment}
This work was supported by JST, ACT-X Grant Number JPMJAX190I, Japan.

%
%
%
%

\end{document}